\pdfoutput=1

\documentclass[11pt]{article}

\usepackage[final]{acl}
\usepackage{amsmath}
\usepackage{times}
\usepackage{latexsym}
\usepackage{booktabs} 
\usepackage{listings}
\usepackage[T1]{fontenc}

\usepackage[utf8]{inputenc}
\usepackage{float}
\usepackage{microtype}

\usepackage{inconsolata}

\usepackage{graphicx}

\title{Swiss Parliaments Corpus Re-Imagined (SPC\_R): 

Enhanced Transcription with RAG-based Correction and Predicted BLEU}

\author{
 \textbf{Vincenzo Timmel\textsuperscript{1}},
 \textbf{Manfred Vogel\textsuperscript{1}},
 \textbf{Daniel Perruchoud\textsuperscript{1}},
 \textbf{Reza Kakooee\textsuperscript{1}}
\\
\\
 \textsuperscript{1}University of Applied Sciences and Arts Northwestern Switzerland \\
 \small{\{vincenzo.timmel, manfred.vogel, daniel.perruchoud, reza.kakooee\}@fhnw.ch}
}

\begin{document}
\maketitle
\begin{abstract}
This paper presents a new long‑form release of the Swiss Parliaments Corpus, converting entire multi‑hour Swiss German debate sessions (each aligned with the official session protocols) into high‑quality speech–text pairs. Our pipeline starts by transcribing all session audio into Standard German using Whisper Large‑v3 under high‑compute settings. We then apply a two‑step GPT‑4o correction process: first, GPT‑4o ingests the raw Whisper output alongside the official protocols to refine misrecognitions, mainly named entities. Second, a separate GPT‑4o pass evaluates each refined segment for semantic completeness. We filter out any segments whose Predicted BLEU score (derived from Whisper’s average token log‑probability) and GPT‑4o evaluation score fall below a certain threshold. The final corpus contains 801 hours of audio, of which 555 hours pass our quality control. Compared to the original sentence level SPC release, our long‑form dataset achieves a 6‑point BLEU improvement, demonstrating the power of combining robust ASR, LLM‑based correction, and data‑driven filtering for low‑resource, domain‑specific speech corpora.

\end{abstract}

\section{Introduction}
Data scarcity in low-resource domains still hinders the development of Automatic Speech Recognition (ASR) systems. For Swiss German, \cite{pluss2021spc} contributed the Swiss Parliaments Corpus (SPC), including a meticulously prepared training dataset with high alignment quality of 176 hours of Swiss German speech paired with Standard German transcripts of Bernese parliamentary debates with a corresponding curated test dataset of 6 hours. The corpus was built using a forced sentence alignment procedure and alignment quality estimator that overcomes challenges such as sentence reordering and language mismatches between Swiss German audio and Standard German text. They used a global alignment algorithm based on Needleman-Wunsch and an Intersection over Union (IoU) estimator to filter out poor-quality alignments. Additional filters, such as character-per-second limits and language detection, ensured that only accurately aligned sentences were included.

The SPC\_R corpus presented in this paper is an extension of the original SPC corpus focusing on the creation, curation, and release of datasets tailored to Swiss German NLP applications. Originally, crawled data from the parliament debates of the Grosser Rat Kanton Bern encompass 801 hours of session recordings in long-form with a length spanning from 28 to 242 minutes paired with official session protocols.

In contrast to \cite{pluss2021spc}, which extracts sentences from parliamentary sessions by finding near-perfect matches between automatically generated transcriptions and the official session protocols, we incorporate an advanced transcription pipeline in SPC\_R. This includes the Whisper Large-v3 model \cite{radford2023robust} for transcription, and a post-correction step using GPT-4o \cite{hurst2024gpt}, aligned with the official protocol to further enhance transcription quality and overall data accuracy.

In addition, the SPC\_R corpus provides the data in long-form, whereas the original SPC is segmented at sentence level.

The primary contributions include:

\begin{itemize}
  \item High-quality transcription by Whisper Large-v3 of approximately 801 hours of audio with high-compute settings, see Section \ref{sec:transcribe_whisper}.
  \item BLEU score \cite{10.3115/1073083.1073135} prediction based on Whisper transcription outputs via linear regression.
  \item A two‑step large language model (LLM) approach in which a first model corrects the transcription and a second, independent model evaluates that correction.
\end{itemize}
This paper provides detailed insights into the methodology, experimental results, and implications for future NLP dataset releases in Swiss German.

\section{Related Work}
\begin{figure}[H]
 \centering
 \includegraphics[width=\columnwidth]{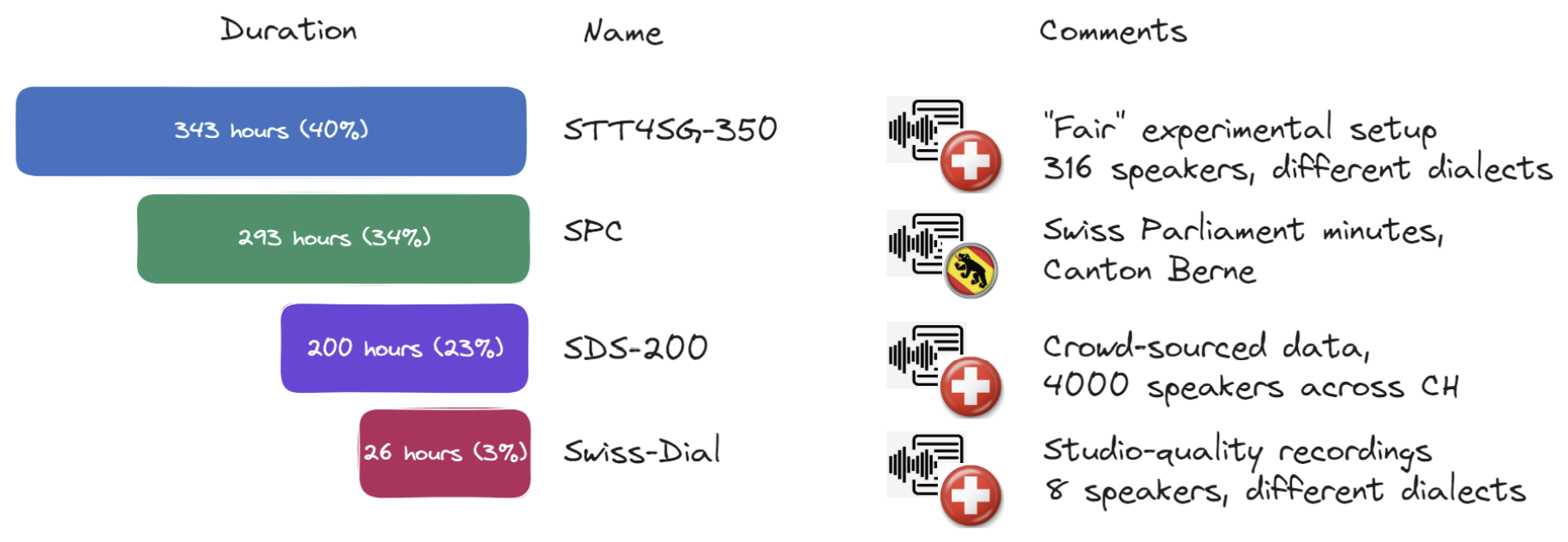}
 \caption{Overview of Swiss German speech to German text datasets. Usage of SPC is possible under MIT license, SDS-200 and STT4SG-350  under SwissNLP license. SwissDial can be used exclusively for research purposes.}
 \label{fig:swiss-stt-data}
\end{figure}

In the past years, several initiatives \cite{pluss2021spc, pluss-etal-2022-sds, pluss2023stt4sg, dogan2021swissdial} made valuable contributions for the development of Swiss German ASR solutions; an overview of the released datasets is shown in Figure \ref{fig:swiss-stt-data}. However, these datasets are all at sentence level which typically does not improve ASR solutions for real-world situations \cite{timmel2024fine}. Additionally, not all existing datasets can be used for commercial purposes.

\section{Transcription with Whisper Large-v3}
\label{sec:transcribe_whisper}
The starting point for the construction of the SPC\_R Corpus is 801 hours of long-form audio from parliament debates of Grosser Rat Kanton Bern which we transcribe with Whisper Large-v3.

Our transcription pipeline uses Whisper Large‑v3 via WhisperX \cite{bain2022whisperx} under high‑compute settings, namely \textit{beam\_size} set to 10, \textit{best\_of} set to 10, and \textit{log\_prob\_threshold} set to –2. All transcriptions are performed on an NVIDIA A4500 GPU with 20 GB of VRAM, using \textit{float16} precision and a \textit{batch\_size} of 8. These high‑compute settings further improve results, as shown in Figure~\ref{fig:WER_improvement}. For all transcribed parliament sessions, we store Whisper's \textit{avg\_log\_prob} output, which reflects the model’s prediction confidence and exhibits strong predictive power for transcription quality, as described in Subsection \ref{sec:bleu_pred}.

\subsection{BLEU Prediction}
\label{sec:bleu_pred}
We observed a linear relationship between the confidence metric calculated by Whisper \cite{kim2023extract}, as presented in Equation \ref{eq:confidence},
and the BLEU score (sacreBLEU\footnote{\url{https://github.com/mjpost/sacreBLEU} (default settings: 4-gram, standard tokenization and smoothing)}, more precisely) of datasets transcribed with Whisper.

\begin{equation}
\begin{aligned}
\text{confidence}
 &= \exp\!\left(\frac{1}{N}\sum_{i=1}^{N} p_i\right)\\[4pt] \label{eq:confidence}
\end{aligned}
\end{equation}

The confidence is derived from Whisper's segment-specific average log-probabilities \textit{avg\_log\_prob}, which are averaged over the whole audio file.
In Equation (\ref{eq:confidence}), 
\(p_i\) denotes the average log-probability for the \(i\)th segment, and \(N\) is the total number of segments in the entire audio file, where a segment is the text between two timestamps predicted by Whisper. Thus, the confidence is the exponential of the average \textit{avg\_log\_prob} over a whole audio file.

\begin{figure}[H]
 \centering
 \includegraphics[width=\columnwidth]{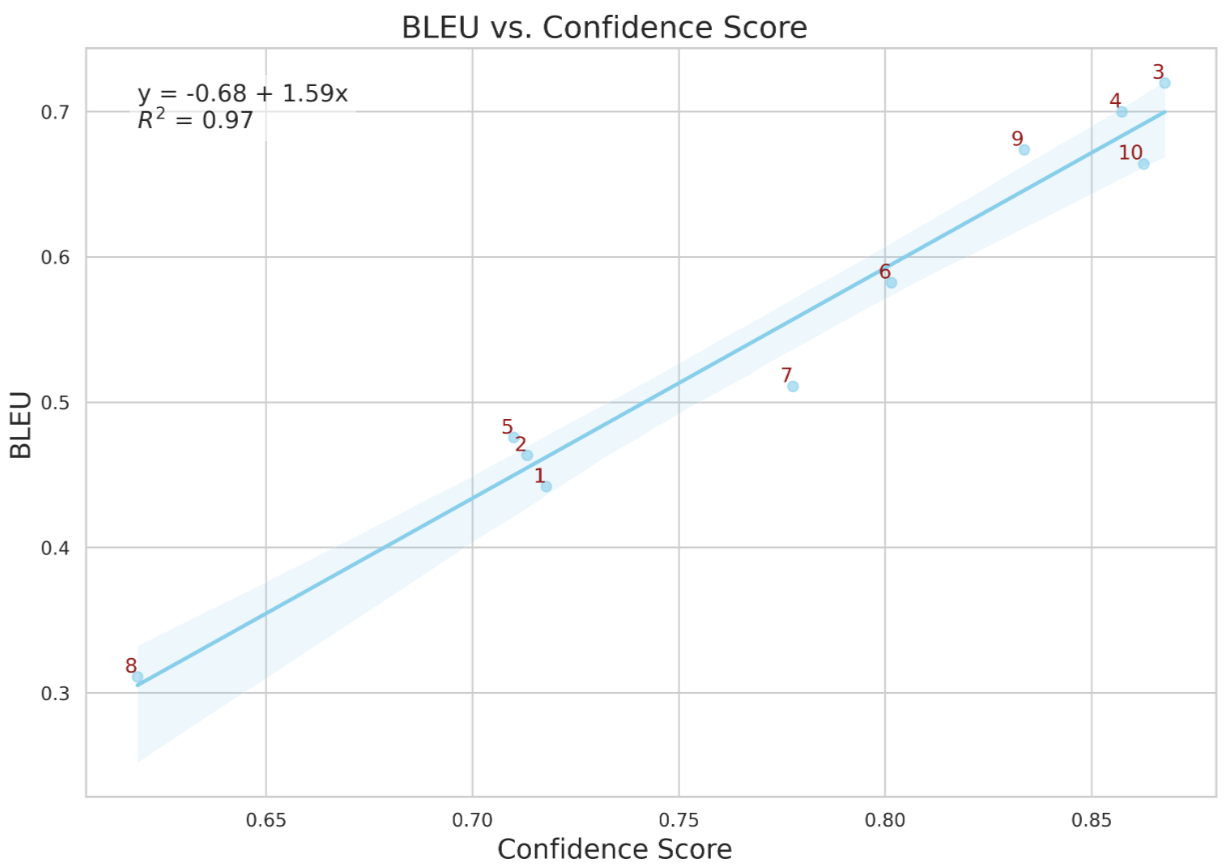}
 \caption{Linear relationship between BLEU score vs. Whisper confidence score for ten long-form conversations, represented by numbers 1-10. The blue shaded area represents the 95\% confidence interval.}
 \label{fig:bleu-vs-confidence}
\end{figure}

Figure \ref{fig:bleu-vs-confidence} shows this linear relationship between the BLEU score (calculated between the transcription and a manually created ground truth) and the confidence on ten distinct, independent  Swiss German datasets. Each dataset of approximately one hour (ca. 8'000 tokens) consists of manually transcribed Swiss German conversations (the ground truth) between two or more speakers (these datasets cannot be disclosed due to data privacy and NDA restrictions). Our analysis shows that higher confidence values are associated with higher BLEU scores in a near-linear fashion, indicating that the confidence metric is a strong predictor of transcription quality, suggesting its potential for assessing transcription performance.

A linear regression fitted to these data produced an intercept of -0.68 and a slope coefficient of 1.59 and allows to predict a BLEU score based solely on the confidence, called the Predicted BLEU, without first creating a ground truth.

\begin{figure}[H]
 \centering
 \includegraphics[width=\columnwidth]{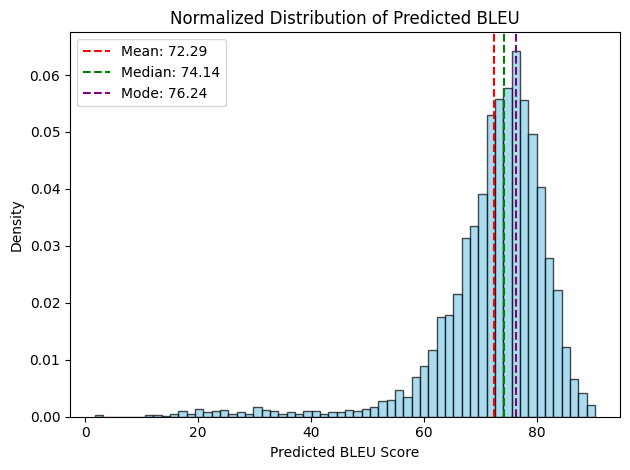}
 \caption{Distribution of Predicted BLEU scores across SPC\_R ($N$ = 131'291 data segments).}
 \label{fig:dist_pred_bleu}
\end{figure}

Figure \ref{fig:dist_pred_bleu} shows the distribution of Predicted BLEU scores for all 131'291 segments of SPC\_R, corresponding to a total of 801 hours of audio.

\begin{figure}[H]
 \centering
 \includegraphics[width=\columnwidth]{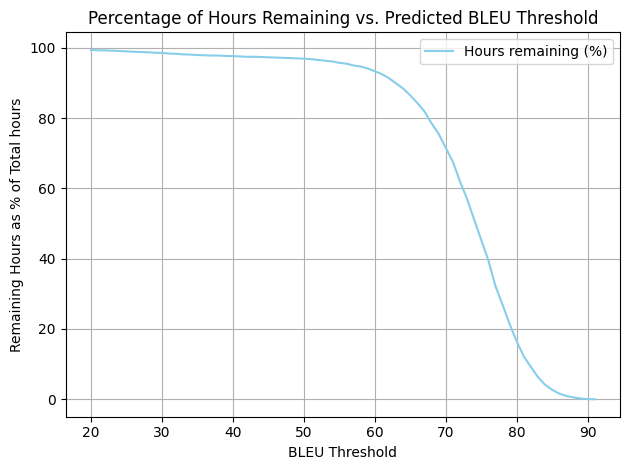}
 \caption{Percentage of data samples that have a BLEU score above the threshold.}
 \label{fig:bleu_threshold}
\end{figure}

Figure \ref{fig:bleu_threshold} shows the cumulative proportion of data samples for a given Predicted BLEU score threshold. As the threshold rises, fewer samples qualify, underscoring the balance between transcription quality and the amount of available data.

Hence, the Predicted BLEU score derived from Whisper’s \textit{avg\_log\_prob} can be used to identify and select high‑quality transcription segments (see Section \ref{sec:sel_data}).

\section{Transcript correction using GPT-4o}

Automated transcription with Whisper Large-v3 shows promising results but leads to errors in named entities (e.g., "Alba Rutschi" instead of "Alberucci") and other similar errors. To mitigate this, we introduce a two-step correction process using text-embedding-3-large GPT-4o and GPT-4o-mini \cite{openai2023embeddings}:
\begin{enumerate}
  \item \textbf{Correction Stage:} GPT-4o is used to refine the initial transcription by prompting it to correct errors, segment by segment. Corrections are based on information injected from the official manual summaries 
 of the parliament session corresponding to the audio segment using Retrieval‑Augmented Generation (RAG, see Subsection \ref{subsec:context_via_rag}).
  \item \textbf{Evaluation Stage:} Evaluation assessments of GPT-4o corrections use manual inspection on small data samples and GPT-4o-mini-as-a-Judge.
\end{enumerate}

GPT-4.1 \cite{openai2025gpt41} was also evaluated but we found that it would repeatedly change conjugation of words, thus sometimes introducing new errors in the transcription. While still overall reducing the WER, it fixed less errors than GPT-4o.

\subsection{Context provision via RAG}
\label{subsec:context_via_rag}
RAG \cite{lewis2020retrieval} is used to provide GPT-4o with factual context to correct the transcription. 

We follow best practices \cite{wang2024searching}, using Faiss \cite{douze2024faiss} for efficient vector storage and retrieval, a sliding window approach and text-embedding-3-large as embedding model. Official manual summaries are ingested with \textit{pyPDF} \cite{pypdf} using chunks of 600 characters with an overlap of 450. These values are chosen to consistently ensure a complete overlap between the transcription and the context from the chunk based on the maximum segment length of 423 characters. We pass the most relevant chunk to GPT-4o as context without re-ranking retrieved chunks. 

Manual evaluation on 122 audio segments corresponding to 50 minutes of transcribed data shows that the correct chunk from the official manual summary is retrieved for 94.1\% of the segments. This high rate may be due to the ease of aligning session protocols with session transcriptions.

\subsection{Correction Stage}
\label{subsec:corr_state}
In the correction stage, GPT-4o is given the context from subsection \ref{subsec:context_via_rag} and the transcription to be corrected, with an extensive, iteratively expanded system prompt specifying usage of the retrieved chunk and additional rules related to peculiarities of the Bernese dialect \footnote{Rules include cases such as "vo dr" (audio) to be corrected from "vor der" to "von der" and "mier" (audio) to be corrected from "mir" to "wir".}. 

\begin{figure}[H]
 \centering
 \includegraphics[width=\columnwidth]{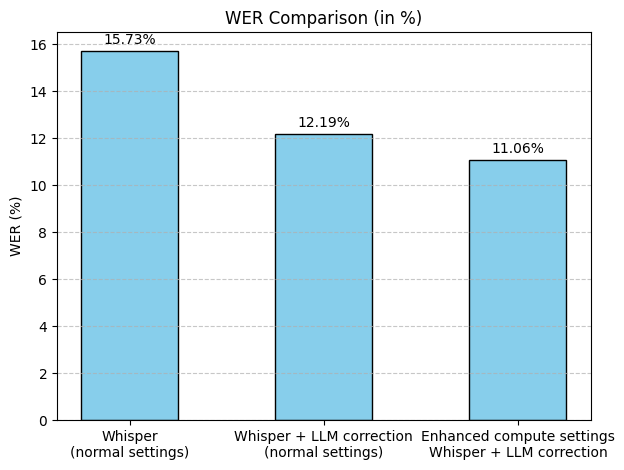}
 \caption{Word Error Rates (WER) for Whisper Large‑v3 under three configurations: standard settings, after applying GPT‑4o correction, and using high-compute settings (enhanced settings) with GPT‑4o correction.}
 \label{fig:WER_improvement}
\end{figure}

The pipeline run with high-compute settings improves the word error rate (WER) from 15.7\% to 11.1\% when evaluated on 50 minutes of manually transcribed data with temperature set to 0.1 to reduce variability and lower WER (see Figure \ref{fig:WER_improvement}).

Additionally, when manually inspecting named entities such as places, names, legal references, and political parties, the correctness of named entity transcriptions increases from initial 72.2\% with Whisper Large-v3 (52 out of 72) to 100\% (72 out of 72) after applying GPT-4o correction. 

Table \ref{tab:correction_answer} shows an example of the audio, the initial Whisper Large-v3 transcription, the context retrieved, and the output corrected with GPT-4o.

\begin{table}[h]
 \caption{Example audio input, initial transcription with Whisper Large-v3, retrieved context (shortened) given to GPT-4o, and its output. GPT-4o is encouraged to keep the correction as close to the input as possible, so that the data can still be used to train an ASR system that relies on aligned audio and text.
}
 \label{tab:correction_answer}
 \centering
 \resizebox{0.9\linewidth}{!}{%
  \begin{tabular}{ l }
 \toprule
 \textbf{\footnotesize{Audio Input} (transcribed)} \\
 \footnotesize{dass ehr au verdaut händ, wenn ehr näbem outo send.} \\
 \midrule
 \textbf{\footnotesize{Whisper Large-v3 output (initial transcription)}} \\
 \footnotesize{dass er auch verdauert hat, wenn er neben dem Auto sitzt.} \\
 \midrule
 \textbf{\footnotesize{Context retrieved via RAG (given to GPT-4o as help for the correction.)}} \\
 \footnotesize{sodass Sie wieder leicht ernüchtert sind und verdaut haben, }\\
  \footnotesize{wenn Sie beim Auto ankommen werden. }\\
  \midrule
   \textbf{\footnotesize{GPT-4o output (final, corrected transcription)}} \\
 \footnotesize{dass Sie auch verdaut haben, wenn Sie neben dem Auto sind.} \\
 \bottomrule
  \end{tabular}%
 }
\end{table}

\subsection{Evaluation Stage}
\label{subsec:eval_stage}
At this stage, the quality of the transcription is evaluated in the following categories (referred to as judgment tokens hereafter):

\begin{itemize} \item \textbf{3) Fully correct:}
All names, nouns, numbers, and abbreviations are accurately transcribed without any mistakes.
\item \textbf{2) Minor error} (not affecting key terms)\textbf{:} 
All names, nouns, numbers, and abbreviations are correct. Small grammatical error present (e.g., incorrect conjugation or article).

\item \textbf{1) Key term error:} 
At least one name, noun, number, or abbreviation is incorrect in the transcription.

\item \textbf{0) No relevant excerpts:} 
The provided excerpt does not contain any relevant content, making evaluation and correction impossible.
\end{itemize}

\begin{figure}[H]
 \centering
 \includegraphics[width=\columnwidth]{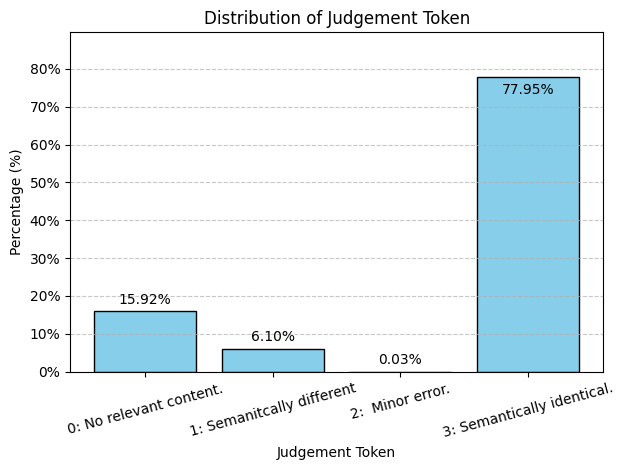}
 \caption{Distribution of the categorization of the final transcription quality using GPT-4o-mini-as-a-judge.}
 \label{fig:judge_dist}
\end{figure}

Figure \ref{fig:judge_dist} presents output of the evaluation stage: 78.0\% of transcripts are semantically identical, which means that the context is perfectly reflected in the transcription, after being corrected by GPT-4o.

After analyzing 50 minutes of data, we discovered that the judgment category is reliable only when we collapse the label “token 0” into “token 1” and likewise merge “token 2” with “token 3.” Grouping the classes this way raises categorization accuracy to 92.2\%. Because GPT-4o-mini struggles to decide whether an error is due to missing context or to a genuine semantic change in the transcription, we fuse those tokens for the final data selection.

\section{Selecting Data and Train/Test Split}
\label{sec:sel_data}
\begin{figure}[H]
 \centering
 \includegraphics[width=0.7\columnwidth]{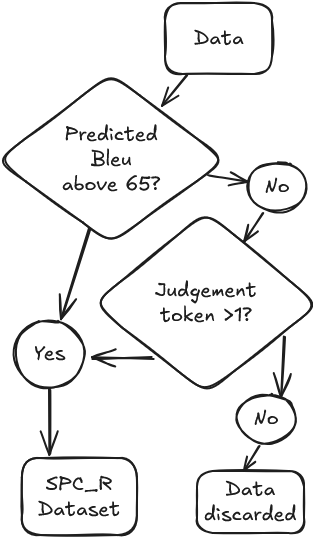}
 \caption{Logic used to build high-quality SPC\_R corpus dataset. Size of initial dataset "Data" is 801 hours of audio, size of high-quality dataset "SPC\_R" is 555 hours.}
 \label{fig:data_flow_spc_r}
\end{figure}

For the construction of the SPC\_R high-quality corpus, we combine findings from Section \ref{sec:bleu_pred} (Predicted BLEU) and Section \ref{subsec:corr_state} (Judgement token) as presented in Figure \ref{fig:data_flow_spc_r}. 

We select a Predicted BLEU score threshold of 65 for filtering based on prior research \cite{GoogleCloudAutoMLEvaluate} suggesting BLEU score above 60 to be indicative 
 of transcription quality superior to general human levels. By choosing a slightly higher threshold, we reduce the variability indicated by the 95\% confidence interval in Figure \ref{fig:bleu-vs-confidence}. While this does not guarantee perfect data, \cite{timmel2024fine} shows that imperfect, pseudo-labelled data can improve the quality of ASR models when used in combination with high-quality training data.

This leads to a high quality corpus of 555 hours of Swiss German audio with paired Standard German transcriptions. For the test set, 30 hours are selected with at least a BLEU score of 70 and segments being evaluated as category 3 (as described in Section \ref{subsec:eval_stage}). The train/test split is therefore 525/30 hours.

\section{Availability and License}
The dataset is publicly available on Hugging Face at \href{https://huggingface.co/datasets/i4ds/spc_r}{i4ds/spc\_r}, the complete codebase (including the prompts) is publicly available on GitHub at \href{https://github.com/i4Ds/spc_r}{i4ds/spc\_r}.

This dataset is released under the Creative Commons Attribution 4.0 International (CC BY 4.0) License, which allows sharing and adaptation provided that appropriate credit is given and any derivatives are licensed under the same terms.\footnote{For more details, see \url{https://creativecommons.org/licenses/by/4.0/}.}

\section{Conclusion}
We present SPC\_R, transcribed with Whisper Large-v3 on high-compute settings, corrected with context by GPT-4o, and evaluated for quality by GPT-4o-mini. This process results in a corpus of 555 hours of high-quality spoken Swiss German paired with Standard German text. 

\section{Future Work}
There are several promising avenues for further enhancing the Swiss Parliaments Corpus. For instance, incorporating additional data sources beyond the Bernese parliamentary debates could broaden the dialectical and contextual diversity of the dataset, potentially leading to performance and robustness improvements of Swiss German ASR models. Exploring alternative transcription models, especially open source solutions, may offer cost or performance advantages over current approaches based on OpenAI models. Finally, there is also room to work with more nuanced evaluation metrics such as Para$_{both}$ \cite{paonessa2023metric}, which better capture semantic fidelity and the accurate transcription of named entities.

\section{Limitations}
\textbf{Evaluation Metrics:}
Our evaluation relies primarily on standard metrics such as BLEU and WER. 
These metrics, while useful, do not capture all aspects of transcription quality, as they can be misleading if a sentence conveys the correct semantics using different words, and especially in terms of correctly transcribing named entities, as they don't weight the greater impact of named entity errors on the comprehension of the transcription. In our experience, most of Whisper's errors, which reduce comprehension of the transcription, are now in the named entities, at least in Swiss German.



\bibliography{custom}

\appendix
\end{document}